\documentclass[fleqn,10pt,twocolumn]{wlscirep}
\usepackage{subcaption}
\usepackage{graphicx}
\usepackage{fontspec}
\usepackage{libertine}
\usepackage{amsmath, amssymb, amsthm, mathrsfs,bm}
\usepackage{booktabs}
\usepackage{url}
\usepackage{graphicx}
\usepackage{subcaption}
\usepackage{enumitem}
\usepackage{algorithm}
\usepackage{algpseudocode}
\usepackage{comment}
\usepackage{multirow}
\usepackage{xcolor}
\title{Using AI to Optimize Patient Transfer and Resource Utilization During Mass-Casualty Incidents: A Simulation Platform}

\author[1, 2]{Zhaoxun ``Lorenz'' Liu}
\author[1, 4]{Wagner H. Souza}
\author[3]{Jay Han}
\author[1,4,†]{Amin Madani}
\affil[1]{Surgical Artificial Intelligence Research Academy, University Health Network, Toronto, Canada}
\affil[2]{Department of Computer Science, University of Toronto, Toronto, Canada}
\affil[3]{Department of Anesthesia and Pain Management, University Health Network, Toronto, Canada}
\affil[4]{Department of Surgery, University of Toronto, Toronto, Canada}
\affil[†]{Corresponding author: amin.madani@uhn.ca}

\begin{abstract}
Mass casualty incidents (MCIs) overwhelm healthcare systems and demand rapid, accurate patient-hospital allocation decisions under extreme pressure. Here, we developed and validated a deep reinforcement learning-based decision-support AI agent to optimize patient transfer decisions during simulated MCIs by balancing patient acuity levels, specialized care requirements, hospital capacities, and transport logistics. To integrate this AI agent, we developed MasTER, a web-accessible command dashboard for MCI management simulations. Through a controlled user study with 30 participants (6 trauma experts and 24 non-experts), we evaluated three interaction approaches with the AI agent (human-only, human-AI collaboration, and AI-only) across 20- and 60-patient MCI scenarios in the Greater Toronto Area. Results demonstrate that increasing AI involvement significantly improves decision quality and consistency. The AI agent outperforms trauma surgeons (p < 0.001) and enables non-experts to achieve expert-level performance when assisted, contrasting sharply with their significantly inferior unassisted performance (p < 0.001). These findings establish the potential for our AI-driven decision support to enhance both MCI preparedness training and real-world emergency response management. 
\end{abstract}
\begin{document}

\flushbottom
\maketitle

\thispagestyle{empty}

\begin{figure*}[t!]
    \centering
    \includegraphics[width=\textwidth]{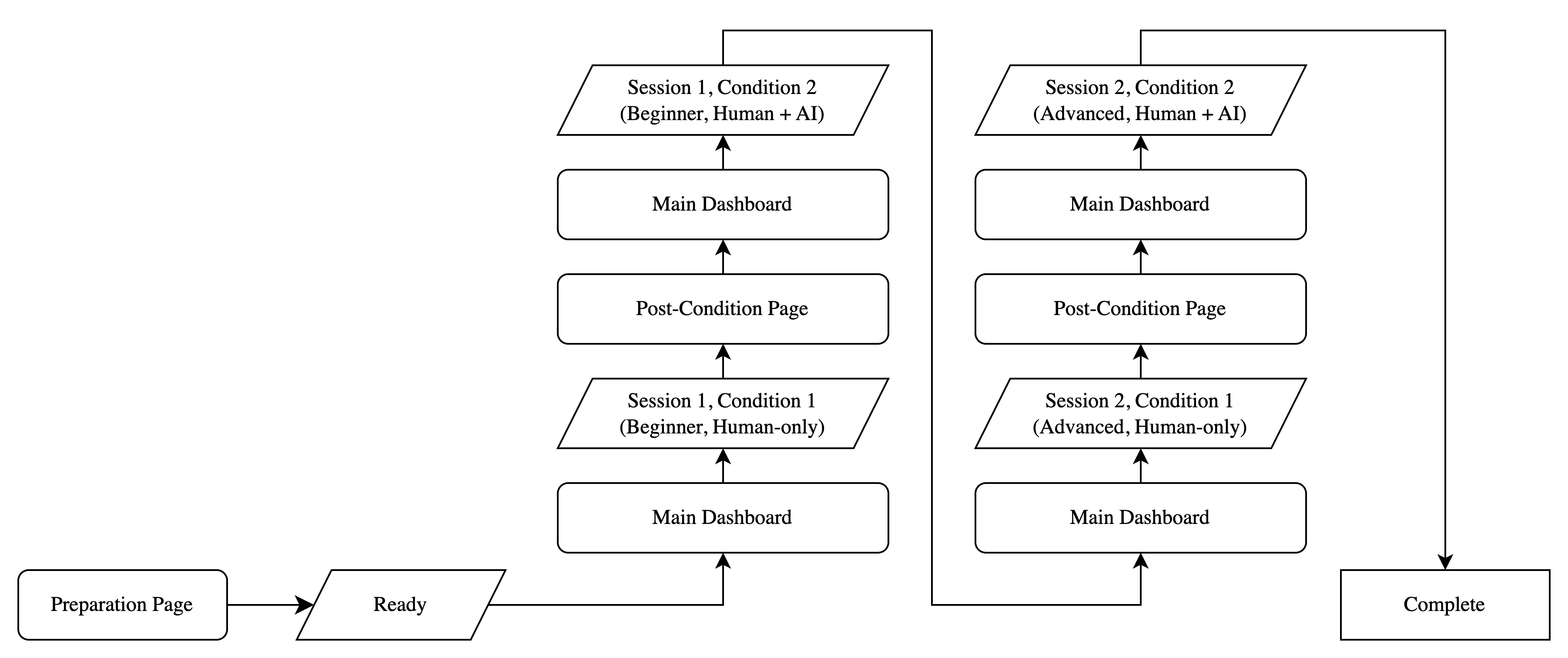}
    \caption{Procedure of the user study.}
    \label{procedure-fig}
\end{figure*}

\section{Introduction}\label{Introduction}

Mass casualty incidents (MCI) are disasters (e.g. natural disasters, explosions, chemical spills, plane crashes, terrorist attacks, military conflict) that overwhelm the local healthcare system and management agencies \cite{bazyar2019triage, bolduc2018comparison}. MCIs require prompt assessment, triaging and transfer of patients to the hospital that is best equipped to accommodate the myriads of potential injuries. Decision-making by MCI commanders can be particularly challenging as they attempt to make critical decisions in a timely fashion while attempting to coordinate with the multiple team members at the disaster site and destination hospitals. This requires detailed understanding of patient factors (e.g. number of victims, mechanisms/types of injuries), hospital factors (e.g. distance from MCI site, available ICU beds/operating rooms/surgeons/ventilators) and transportation factors (e.g. available ambulances, helicopters) to optimize transfer decisions and patient outcomes. Given the relative rarity and uniqueness of each MCI, training healthcare workers in MCI response decision-making is limited mostly to tabletop simulation exercises \cite{tahernejad2024application, albahri2024systematic}. There is therefore a need for innovative methodologies for improving decision-making for MCI in a cost-effective and scalable manner. One potential solution is through the use of digital solutions and intelligent systems for providing end-users with simulation-based training and decision-support to cope with these challenges (\textbf{C}s): 

\begin{itemize}
    \setlength{\itemsep}{0pt} 
    \item \textbf{C1}. Uncertainty, Cognitive Overload, and Inefficiency: Commanders lack real-time visibility into total casualty numbers and conditions across the scene. Additionally, the quantity of patients could overwhelm judgment, which could hamper optimal resource deployment and increase the likelihood of error-prone decisions \cite{bazyar2022accuracy}.
    \item \textbf{C2}. Environmental Vulnerabilities: Paper-based tools are susceptible to weather conditions and physical damage.
    \item \textbf{C3}. Update and Tracking Difficulties: Communicating and tracking changes in patients and hospitals across the response team is slow and troublesome \cite{khorram2023implication}.
    \item \textbf{C4}. Lost Education \& Training Opportunities: Valuable data about response patterns and decision-making is not captured for future analysis.
\end{itemize}

Reinforcement Learning (RL) \cite{reinforcement-survey, sutton-reinforcement} is a machine learning paradigm that enables agents to learn optimal behaviors through trial-and-error interactions with complex simulated environments while maximizing long-term rewards. RL proves particularly valuable when structured training data is scarce, and this is exactly the situation for historical MCIs. Deep Learning (DL) \cite{LeCun-deep-learning, ian-deep} utilizes multi-layered artificial neural networks for feature extraction and prediction generation, which is extremely suitable for MCIs because of their high-dimensional information. Deep Reinforcement Learning (DRL) \cite{drl-survey}, on the other hand, integrates the two approaches, combining their respective strengths, and can continuously adapt and learn in complex environments \cite{mnih2016asynchronous, chen2022syntheticdata, baucum2021improvingdrl}. Moreover, Proximal Policy Optimization (PPO) \cite{schulman2017proximal}, a DRL algorithm, provides a promising approach. The stability and sample efficiency of PPO make it especially appropriate for MCI contexts where suboptimal decisions carry significant consequences. Prior to this work, DRL has emerged as a powerful tool for decision-making in healthcare \cite{mousavi2018deep, yu2019reinforcement}. In healthcare transport, we have seen applications in ambulance dispatch \cite{gandhi2020applications, greco2021artificial}. Other applications include dynamic treatment regimes \cite{ji2019deep, prasad2017reinforcement}, emergency department management \cite{lee2020improving, liu2018deep}, process control \cite{gandhi2020applications, mahmood2018benchmarking}, drug discovery \cite{mousavi2018deep, zhavoronkov2019deep}, and personalized health recommendations \cite{gandhi2020applications, yauney2018reinforcement}.

Therefore, we aimed to 1) develop a novel digital platform to simulate MCI events, 2) train and validate an AI algorithm with DRL that provides decision-support to accelerate and optimize patient transfer decision, and 3) determine whether it improves decisions amongst both trauma experts and non-experts in a simulated environment. 

\begin{figure*}[t!]
    \centering
    \begin{subfigure}[b]{0.3\textwidth}
        \centering
        \includegraphics[width=\textwidth]{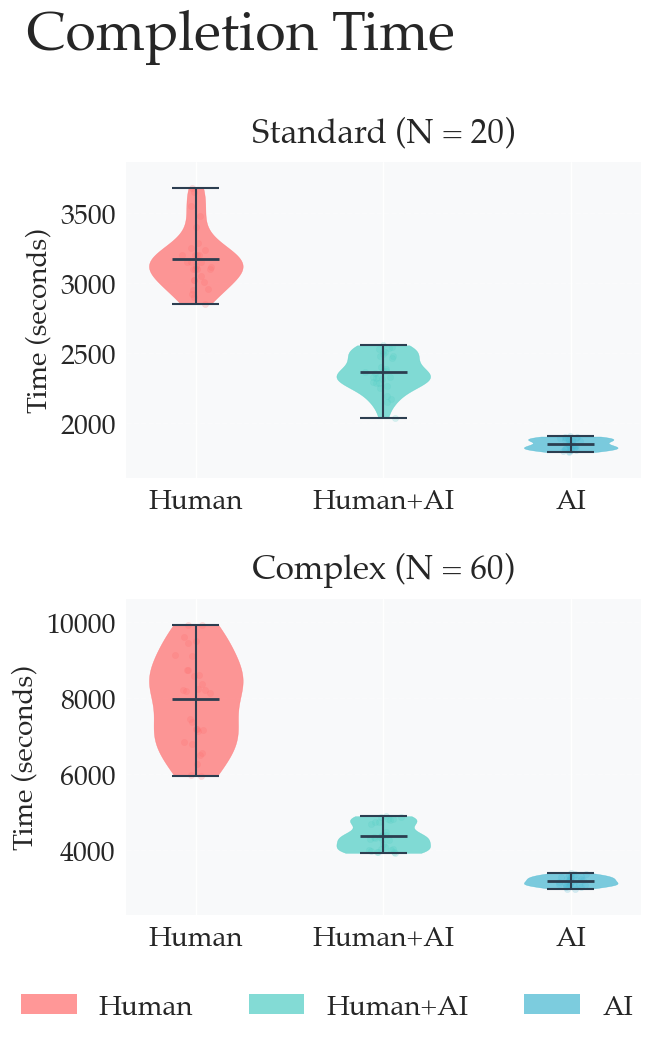}
        \caption{Completion Time (s)}
        \label{fig:subfig1}
    \end{subfigure}
    \hfill
    \begin{subfigure}[b]{0.3\textwidth}
        \centering
        \includegraphics[width=\textwidth]{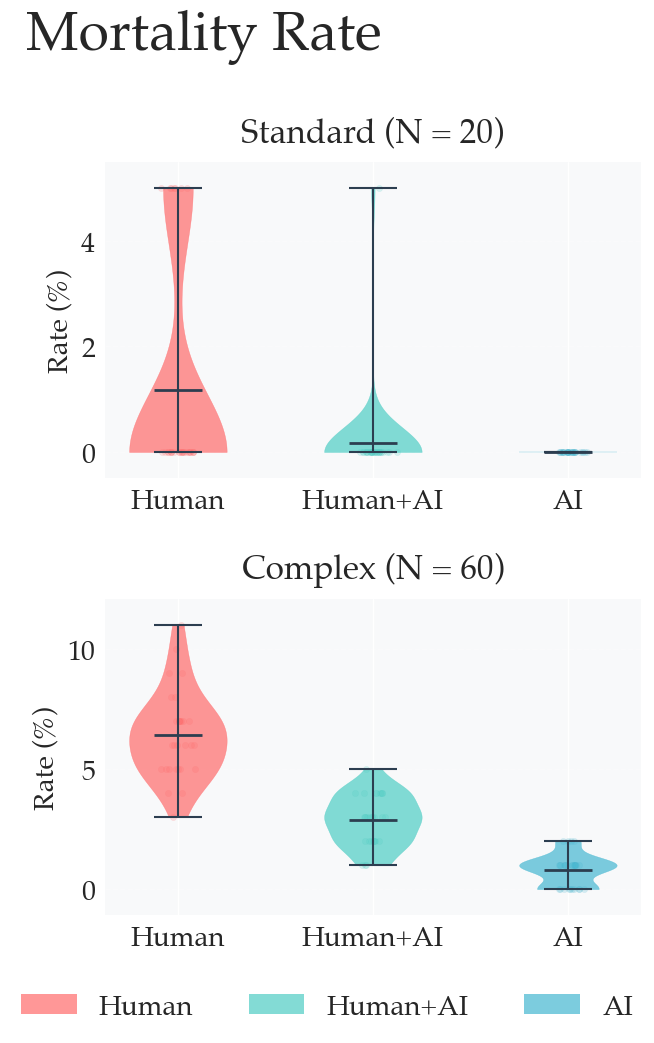}
        \caption{Mortality Rate (\%)}
        \label{fig:subfig2}
    \end{subfigure}
    \hfill
    \begin{subfigure}[b]{0.3\textwidth}
        \centering
        \includegraphics[width=\textwidth]{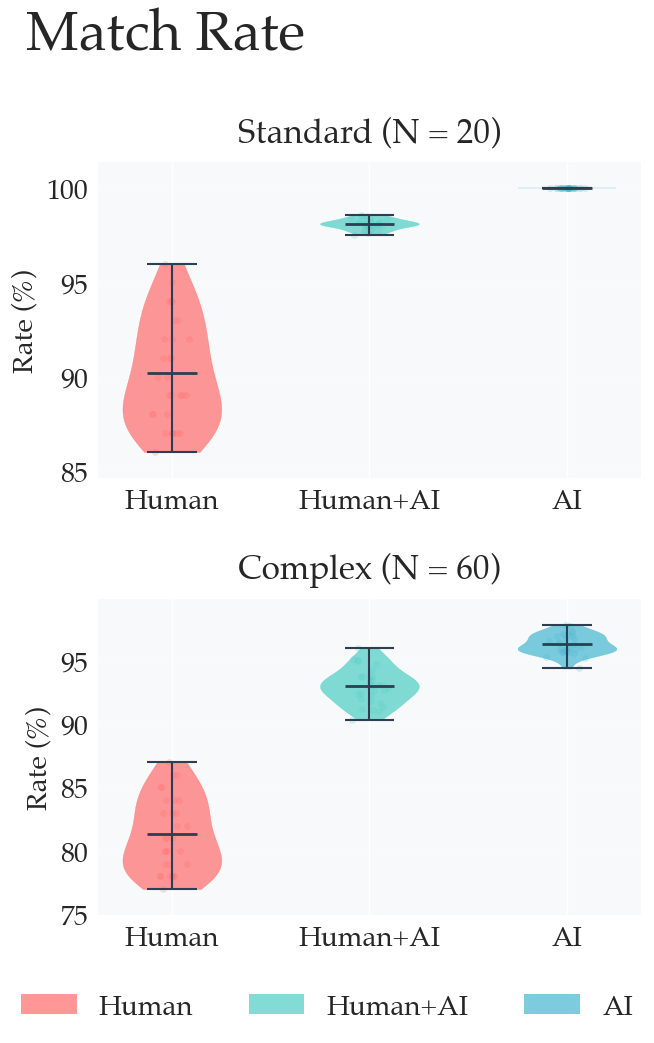}
        \caption{Match Rate (\%)}
        \label{fig:subfig3}
    \end{subfigure}
    \caption{Quantitative results from the user study. For Completion Time and Mortality Rate, lower values indicate better performance, while for Match Rate, higher values indicate better performance. }
    \label{fig:user-study-res}
\end{figure*}

\section{Results}\label{Results}

\subsection{Study Design}

To evaluate whether MasTER facilitates MCI management, we conducted user studies using two distinct simulation exercises: a Standard level (20 patients) and Complex level (60 patients), both in the Greater Toronto Area. Users participated in both simulation in two iterations: 1) Human-only (no AI available to assist) and 2) Human+AI (human-in-the-loop approach where AI assistance was available at their discretion). In the Human+AI setting, participants could request AI-generated suggestions for patient assignments which they could either accept or decline. We also tested the AI model as a standalone fully autonomous agent making decisions (AI-only). Participants (including trauma surgeons and non-trauma surgeons) were recruited to complete the Human-only and Human+AI simulations. Prior to starting the simulation exercises, all participants underwent a training module to gain familiarity with the platform. 

The procedure of the user study is shown in Figure \ref{procedure-fig}. The study began with a standardized training session in the preparation page to minimize learning effects. Each participant then completed two sessions:

\begin{enumerate}
    \item Session 1 (Standard Level) consisted of:
\begin{itemize}[nosep]
   \item Initial condition (Human-only or Human+AI, counterbalanced)
   \item 2-minute rest period in the Post-Condition Page
   \item Second condition
   \item Post-session questionnaires: NASA Task Load Index (NASA-TLX) \cite{hart1986nasa} and System Usability Scale (SUS) \cite{sus}
   \item Self-paced rest period
\end{itemize}
\item Session 2 (Complex Level) followed an identical structure but with increased complexity (60 patients). Both sessions were designed to evaluate decision-making performance under varying cognitive loads while maintaining systematic data collection throughout the experimental procedure. 

\end{enumerate}

Specifically, through the user study, we aimed to answer the following research questions (\textbf{RQ}s):

\begin{itemize}
    \setlength{\itemsep}{0pt} 
    \item \textbf{RQ1}. Whether MasTER improves decisions amongst both trauma experts and non-experts in a simulated environment?
    \item \textbf{RQ2}. What is the perceived utility of MasTER's features?
    \item \textbf{RQ3}. Do participants consider MasTER as a useful, robust, and efficient tool to use in managing MCIs?
\end{itemize}

\subsection{Participants}

We recruited participants (N = 30) using two sampling approaches. First, purposive sampling established our initial participant group (N = 6) from within the professional trauma community; this group forms our expert participants. Second, snowball sampling yielded our remaining participants (N = 24) from the student community; this group forms our non-expert participants. 

Participants from purposive sampling are all experienced trauma professionals ($M_{age}$=41.12, $SD$=6.12, 4 males and 2 females); participants from snowball sampling are medical trainees ($M_{age}$=25.82, $SD$=3.14, 13 males and 11 females). All participants provided informed consent through REDCap \cite{redcap1, redcap2} following institutional review board approval. 

\subsection{Metrics}

Our evaluation framework incorporated both quantitative and qualitative measures to provide a comprehensive assessment of the MasTER platform. The system automatically collected quantitative metrics including:

\begin{enumerate}[noitemsep]
\item Total completion time: $T_{\text{end}} - T_{\text{start}}$, where $T_{\text{start}}$ is when the simulation begins and $T_{\text{end}}$ is when the final decision is made.

\item Patient mortality rate: $\frac{N_{\text{deaths}}}{N_{\text{total}}} \times 100\%$, where $N_{\text{deaths}}$ is the number of patients who died during the simulation and $N_{\text{total}}$ is the total number of patients.

\item Resource match rate: $R_{\text{match}} = \frac{1}{P} \sum_{n=1}^{P} \frac{N_{\text{correct}_n}}{N_{\text{total}_n}} \times 100\%$, where $N_{\text{correct}_n}$ is the number of matched resources for the $n$-th patient, $N_{\text{total}_n}$ is the total number of needed resources for the $n$-th patient, and $P$ is the total number of patients.
\end{enumerate}

For qualitative assessment, we measured workload using the NASA Task Load Index (NASA-TLX) \cite{nasatlx} and system usability via the System Usability Scale (SUS) \cite{sus}. We also assessed user perception on the utility and value of the tool. These post-condition evaluations captured immediate impressions and experiences while minimizing recall bias. All questionnaires were administered and collected via REDCap \cite{redcap1, redcap2}.

\begin{table*}[t!]
\centering
\caption{Expert vs. Non-Expert Mean Performance Comparison Across All Metrics}
\label{tab:expert-nonexpert-all}
\begin{tabular}{llcccc}
\toprule
\multirow{2}{*}{\textbf{Metric}} & \multirow{2}{*}{\textbf{Level}} & \multicolumn{2}{c}{\textbf{Human-only}} & \multicolumn{2}{c}{\textbf{Human+AI}} \\
\cmidrule(lr){3-4} \cmidrule(lr){5-6}
& & \textbf{Experts} & \textbf{Non-Experts} & \textbf{Experts} & \textbf{Non-Experts} \\
\midrule
\multirow{2}{*}{Completion Time (s)} & Standard & 3136.17 (35.66) & 3179.42 (227.59) & 2358.33 (70.45) & 2363.54 (139.85) \\
& Complex & 6342.00 (301.24) & 8375.83 (923.19) & 4386.50 (315.25) & 4346.88 (333.49) \\
\midrule
\multirow{2}{*}{Mortality Rate (\%)} & Standard & 0.00 (0.00) & 1.46 (2.21) & 0.00 (0.00) & 0.21 (1.02) \\
& Complex & 4.50 (1.97) & 6.92 (1.38) & 3.00 (1.10) & 2.88 (1.08) \\
\midrule
\multirow{2}{*}{Match Rate (\%)} & Standard & 92.00 (3.52) & 89.67 (2.46) & 98.13 (0.42) & 98.07 (0.28) \\
& Complex & 84.67 (3.67) & 80.50 (2.30) & 93.61 (1.58) & 92.79 (1.42) \\
\bottomrule
\multicolumn{6}{p{\textwidth}}{\small Note: Values in parentheses represent standard deviations.} \\
\end{tabular}
\end{table*}

\subsection{Analysis}

There was a total of 30 participants, including 6 expert trauma surgeons from a high-volume Level 1 trauma hospital, and 24 non-experts. For the entire cohort, there were significant differences across conditions (Human-only, Human+AI, and AI-only) for all performance metrics. For completion time (Figure \ref{fig:subfig1}), repeated measures ANOVA showed a significant main effect (F(2,87) = 892.31, p < .001, η²p = .943). Post-hoc Tukey's HSD tests indicated significant differences between all pairs of conditions (p < .001). The Human+AI condition demonstrated substantially improved performance compared to the Human-only condition (d = 4.72, 95\% CI [4.23, 5.21]). 

In the Standard level, Human+AI participants completed tasks 25.49\% faster than Human-only (Human: M = 3170.77s, SD = 204.92s; Human+AI: M = 2362.50s, SD = 129.01s). This improvement was even more pronounced in the Complex level with a 45.35\% reduction in completion time (Human: M = 7969.07s, SD = 1166.96s; Human+AI: M = 4354.80s, SD = 330.30s).

There was a significant improvement in simulated mortality rates (Figure \ref{fig:subfig2}), using AI assistance for the Standard level (t(29) = 7.82, p < .001; Human: M = 1.17\%, SD = 2.11\%; Human+AI: M = 0.17\%, SD = 0.90\%), representing an 85.71\% reduction in mortality. This improvement was similarly pronounced during Complex scenarios where there were more patients to be triaged (t(29) = 11.23, p < .001; Human: M = 6.43\%, SD = 1.76\%; Human+AI: M = 2.87\%, SD = 1.02\%), showing a 55.44\% reduction in mortality.

Match rates (Figure \ref{fig:subfig3}) similarly improved with AI assistance, with the Standard level achieving near-perfect scores in the Human+AI condition (M = 98.09\%, SD = 0.27\%) compared to Human-only (M = 90.17\%, SD = 2.76\%; t(29) = 15.82, p < .001), an 8.78\% improvement. In the Complex level, the Human+AI condition (M = 92.98\%, SD = 1.36\%) significantly outperformed the Human-only condition (M = 81.30\%, SD = 2.84\%; t(29) = 21.69, p < .001), representing a 14.36\% improvement in match rate.

When analyzing expert versus non-expert performance, we found that non-experts were unable to match experts' performance in the Human-only condition across all three metrics. For completion time, experts were faster in both Standard (Experts: M = 3136.17s, SD = 35.66s; Non-experts: M = 3179.42s, SD = 227.59s) and Complex levels (Experts: M = 6342.00s, SD = 301.24s; Non-experts: M = 8375.83s, SD = 923.19s). Mortality rates were lower for experts in both Standard (Experts: M = 0.00\%, SD = 0.00\%; Non-experts: M = 1.46\%, SD = 2.21\%) and Complex levels (Experts: M = 4.50\%, SD = 1.97\%; Non-experts: M = 6.92\%, SD = 1.38\%). Match rates were also higher for experts in both Standard (Experts: M = 92.00\%, SD = 3.52\%; Non-experts: M = 89.67\%, SD = 2.46\%) and Complex scenarios (Experts: M = 84.67\%, SD = 3.67\%; Non-experts: M = 80.50\%, SD = 2.30\%). However, with AI assistance, non-experts demonstrated remarkable improvement across all metrics. For completion time in Complex scenarios, non-experts with AI (M = 4346.88s, SD = 333.49s) outperformed experts without AI (M = 6342.00s, SD = 301.24s). Non-experts' mortality rates with AI (Complex: M = 2.88\%, SD = 1.08\%) were lower than experts without AI (M = 4.50\%, SD = 1.97\%). Similarly, non-experts' match rates with AI (Complex: M = 92.79\%, SD = 1.42\%) exceeded those of unaided experts (M = 84.67\%, SD = 3.67\%).

Effect sizes for the improvements were particularly large for completion time (Standard level: d = 4.72; Complex level: d = 4.21) and match rates (Standard level: d = 4.04; Complex level: d = 5.24), highlighting the substantial practical significance of the AI assistance. Mortality rate reductions also showed meaningful effect sizes, particularly in the Complex scenarios (Standard level: d = 0.62; Complex level: d = 2.47).

Qualitative analysis of NASA-TLX scores indicated significantly lower perceived workload in the Human+AI condition (M = 31.40, SD = 3.78) compared to Human-only (M = 63.7, SD = 7.2; t(29) = 17.92, p < .001). The System Usability Scale (SUS) score for the Human+AI system was exceptional at 87.87 (SD = 2.50), placing it in the 95th percentile of evaluated systems.

\section{Discussion}\label{Discussion}

We organize this section according to our previously proposed research questions (\textbf{RQ}s in Section \ref{Results}).

\textbf{RQ1}: In simulated MCI, the use of the AI model significantly improved triage decisions, with substantial mortality rate reductions (85.71\% for the Standard level, 55.44\% for the Complex level; p < .001). The large effect size for completion time (d = 4.72) shows both statistical and practical significance. The DRL model performed even better autonomously than with human intervention, suggesting that while human expertise remains valuable, there may be instances where algorithmic decision-making outperforms human-in-the-loop approaches in time-critical scenarios. Our findings align with recent studies showing AI systems can achieve expert-level performance in emergency medicine tasks with complex, multi-variable inputs. The significant mortality reductions across both complexity levels demonstrate the model's robustness and potential generalizability.

The novel application of DRL to MCI management represents a significant innovation in emergency medicine, offering a scalable and cost-effective training solution that addresses known limitations in MCI preparedness. However, we acknowledge that our validation relies on simulated data rather than real-world implementation. Our ongoing exploration will help bridge this gap, though additional validation across different healthcare systems, regions, and resource availability scenarios will be necessary to establish broader applicability. Future work should include pilot implementations in hospital systems to evaluate real-world efficacy and interoperability with existing emergency response software and electronic health records (EHRs).

\textbf{RQ2}: MasTER's features demonstrated strong utility, with NASA-TLX scores showing 50.7\% reduced workload (p < .001) and exceptional usability (SUS: 87.87, 95th percentile). Features provided crucial support in complex scenarios, particularly in the Complex level where there are many victims and decisions become extremely challenging. The substantial workload reduction is particularly noteworthy given the high-stress, cognitive-heavy nature of MCI management. This reduction occurred without compromising decision quality, suggesting MasTER effectively offloads cognitive burden while maintaining or improving performance standards. The exceptionally high SUS score exceeds industry benchmarks for healthcare technologies, indicating potential for rapid adoption and minimal training requirements. The interface design principles employed in MasTER could inform future emergency management systems, particularly the visualization components that reduced information overload during complex scenarios.

While our comprehensive evaluation metrics incorporate both quantitative and qualitative measures, providing a well-rounded assessment of system effectiveness, we recognize limitations in algorithm transparency and explainability. Future iterations should include feature importance analysis to better understand which variables (e.g., travel time, hospital resources, patient severity) most influenced the AI's decision-making process. This would enhance trust and adoption among healthcare professionals who may be hesitant to rely on "black-box" algorithms for critical decisions. Additionally, further research is needed to clarify how decision accountability is distributed between AI and human users, particularly in ambiguous situations where AI recommendations might conflict with expert judgment.

\textbf{RQ3}: Participants generally found MasTER useful, robust, and efficient, as demonstrated by high satisfaction scores and significant time improvements (25.49\% faster for the Standard level, 45.35\% for the Complex level; p < .001). The greater relative improvement in complex scenarios suggests MasTER provides exponential benefits as cognitive demands increase. Qualitative feedback from trauma experts highlighted the system's potential to standardize care approaches across different experience levels, potentially addressing disparities in emergency response capabilities between resource-rich and resource-limited settings. The time improvements observed are clinically meaningful in the context of the ``golden hour" principle in trauma care, where minutes saved correlate directly with survival outcomes.

The positive reception among domain experts suggests potential for implementation, though further field testing in actual MCI scenarios would be necessary to validate these findings beyond simulation environments. Future iterations should address the suggested enhancements, such as incorporating neurosurgical availability and helicopter transportation options, with a clear timeline and feasibility assessment for these improvements. We also recognize the importance of addressing potential regulatory and logistical challenges in deploying AI-driven decision-support tools in emergency settings. Longitudinal studies assessing actual patient and system outcomes over time will be essential to fully evaluate MasTER's impact on emergency response and patient survival rates in real-world scenarios.

Our human-in-the-loop approach balances AI capability with human expertise, allowing for real-world applicability while demonstrating AI's potential to augment decision-making. However, potential biases in synthetic training data, particularly regarding differences in healthcare infrastructure across regions, must be carefully addressed in future work. As we move forward, we will focus on critical factors for real-world application, such as real-time data ingestion capabilities and linkage to hospital databases for accurate resource availability.

\begin{figure*}[t!]
    \begin{subfigure}{\linewidth}
        \includegraphics[width=\linewidth]{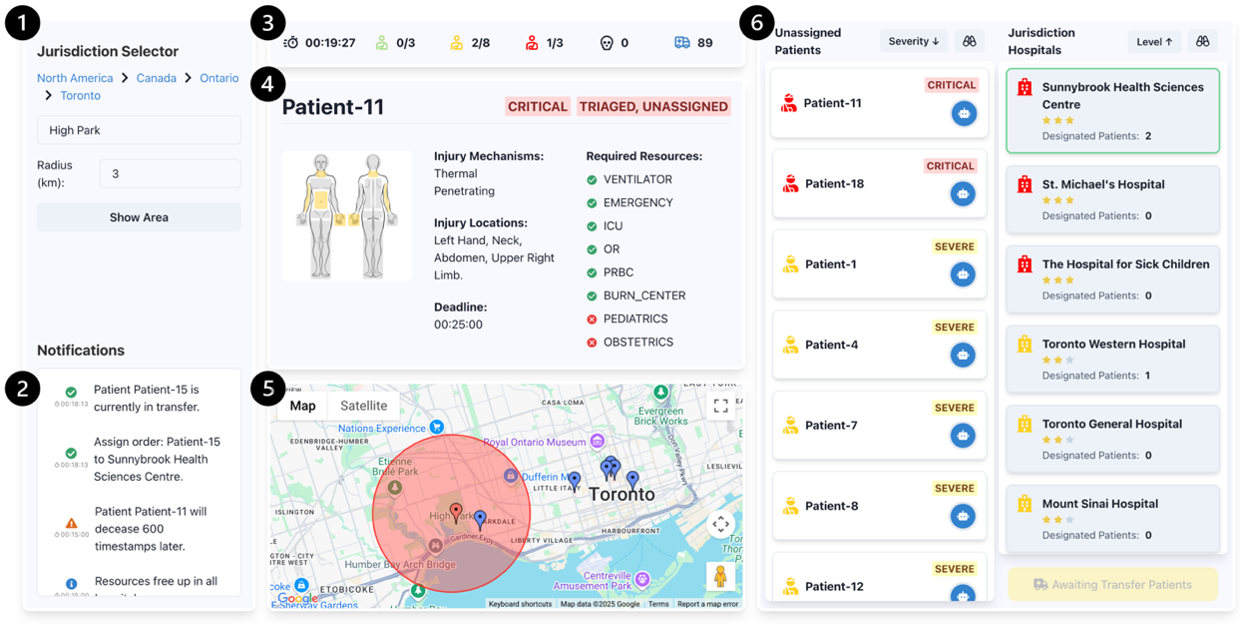}
        \caption{The main user interface of MasTER. Each component's detailed explanation is in Section \ref{MasTER}}
        \label{figure1-top}
    \end{subfigure}
    
    \vspace{1em} 
    
    \begin{subfigure}{\linewidth}
        \includegraphics[width=\linewidth]{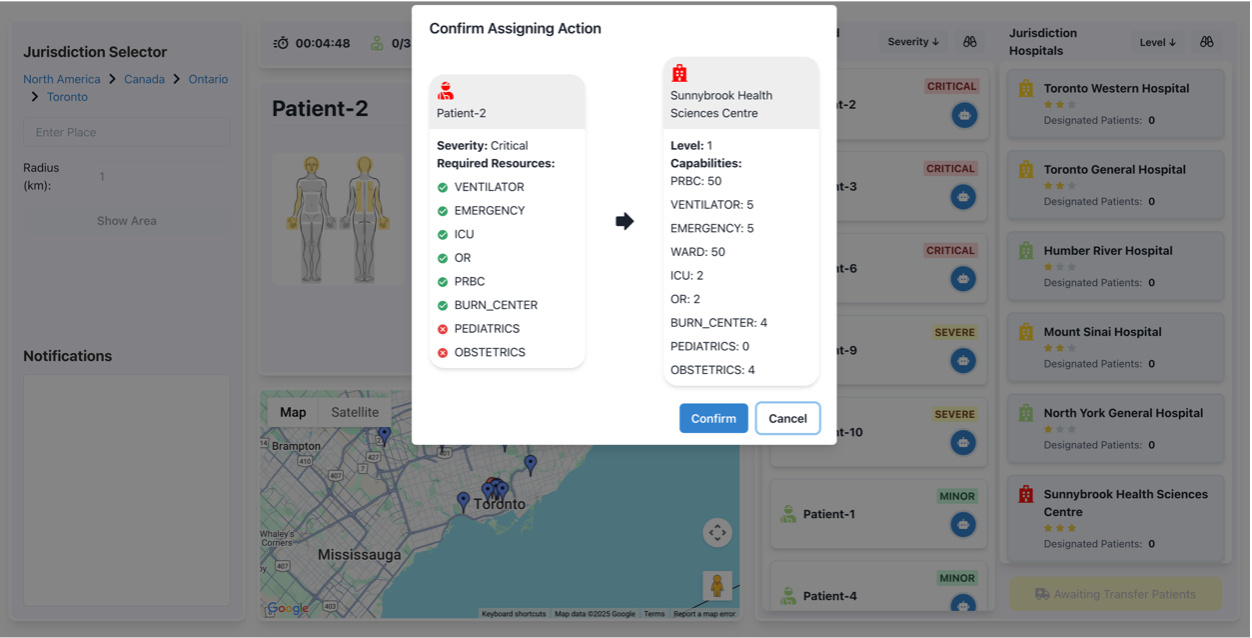}
        \caption{An example of AI suggestion. }
        \label{figure1-bottom}
    \end{subfigure}
    
    \caption{MasTER: an intelligent human-in-the-loop command dashboard. }
    \label{fig:mainfigure}
\end{figure*}

\begin{table*}[ht]
\centering
\caption{Mapping of MasTER Components to Challenges Addressed (Stated in Section \ref{Introduction})}
\label{tab:component-challenges}
\begin{tabular}{p{4cm}p{10cm}}
\hline
\textbf{Component} & \textbf{Challenges Addressed} \\
\hline
\textit{Jurisdiction Selector} & 
\textbf{C1}: Reduces cognitive overload by enabling focused management of geographical areas.
\newline \textbf{C3}: Facilitates efficient updates within specific jurisdictions, improving communication.\\
\hline
\textit{Notifications} & 
\textbf{C1}: Mitigates uncertainty through color-coded, timestamped updates.
\newline \textbf{C3}: Provides rapid communication about changes in patient status and hospital availability.
\newline \textbf{C4}: Creates a log of events that can be used for post-incident analysis and training.\\
\hline
\textit{Status Bar} & 
\textbf{C1}: Reduces cognitive overload by displaying critical metrics in real-time, enhancing situational awareness.
\newline \textbf{C4}: Captures key metrics throughout the incident for later analysis and training.\\
\hline
\textit{Detail Panel} & 
\textbf{C1}: Mitigates uncertainty by providing comprehensive patient and hospital information.
\newline \textbf{C2}: Digital format eliminates vulnerability to environmental conditions.
\newline \textbf{C3}: Enables rapid access to updated information about patients and hospitals.\\
\hline
\textit{Interactive Map} & 
\textbf{C1}: Reduces cognitive overload through spatial visualization, improving decision efficiency.
\newline \textbf{C2}: Digital format protects against environmental damage.
\newline \textbf{C4}: Spatial patterns can be recorded for future training purposes.\\
\hline
\textit{Draggable Action Panel} & 
\textbf{C1}: Directly addresses inefficiency by streamlining assignment process and reducing error-prone decisions through AI suggestions.
\newline \textbf{C2}: Eliminates paper-based vulnerabilities.
\newline \textbf{C3}: Enables real-time tracking of patient assignments and hospital capacity.
\newline \textbf{C4}: Records assignment decisions for post-incident review and education.\\
\hline
\end{tabular}
\end{table*}

\section{Methods}\label{Methods}

\subsection{Deep Reinforcement Learning Agent}

To streamline and support patient-hospital transfer decision-making, an AI agent was trained with DRL, specifically, the algorithm of PPO. The AI agent was trained on extensive simulated MCI scenarios (n=10,000), representing diverse casualty volumes (ranging from 10-500 patients), injury patterns, and regional hospital resource configurations. The modelling of simulated MCIs strictly follows the existing rapid trauma triage protocols, such as color codes for patient severity and levels for hospital capability \cite{clarkson2024ems}. At its core, the agent optimized a multi-objective reward function prioritizing survival probability while considering transport duration, facility capacity constraints, and specialty care requirements. Once the model was developed, it was integrated into the simulation platform as an add-on feature where end-users can summon the model to provide a recommendation on the optimal destination hospital for any given patient.


\subsubsection{Training Environment \& State Space}
The environment for the DRL training represents an MCI scenario with:
\begin{itemize}
    \item \textbf{Patient State Space}: Each patient $p_i \in P$ is defined as $p_i = (id_i, s_i, r_i)$ where:
        \begin{itemize}
            \item $id_i$ is a unique identifier
            \item $s_i \in \{0,1,2,3\}$ represents severity according to the triage code (0: deceased, 1: minor, 2: severe, 3: critical) \cite{clarkson2024ems}
            \item $r_i \in \{0,1\}^8$ is a binary requirements vector for needed medical resources, including ventilator, emergency, ICU, operating room (OR), pRBC, burn center, pediatrics, and obstetrics. 
        \end{itemize}
    \item \textbf{Hospital State Space}: Each hospital $h_j \in H$ is defined as $h_j = (id_j, l_j, \lambda_j, c_j)$ where:
        \begin{itemize}
            \item $id_j$ is a unique identifier
            \item $l_j \in \mathbb{R}^2$ represents geographical coordinates
            \item $\lambda_j \in \{1,2,3\}$ denotes hospital level
            \item $c_j \in \mathbb{N}^8$ represents resource capacities, including ventilator, emergency, ICU, operating room (OR), pRBC, burn center, pediatrics, and obstetrics. 
        \end{itemize}
    \item \textbf{Travel Duration Matrix}: A matrix $D \in \mathbb{R}_{\geq0}^{|P|\times|H|}$ with elements $D_{ij} = D(L,l_j)$ representing travel times from incident location $L$ to each hospital
\end{itemize}

\subsubsection{Action Space}
The action space $A$ at time $t$ is defined as:
\begin{equation}
A_t = \{a \in \{1,...,|H|\}^{|P|} : C(a) \leq c\}
\end{equation}
where $a = [a^1,...,a^{|P|}]$ represents hospital assignments, $C(\cdot)$ is the capacity utilization function, and $c$ represents hospital capacity constraints.

\subsubsection{Reward Function}
The reward function in MasTER is designed as a comprehensive linear model that balances multiple objectives based on patient severity, resource availability, and time-critical factors. For a given state-action pair, the total reward $R$ is:

\begin{equation}
R = \sum_{i \in P} R_i
\end{equation}

For surviving patients, we define two critical penalty factors:

\begin{itemize}
    \item Time-sensitive penalty $PT_i \in [0, 1]$:
    \begin{equation}
    PT_i = \max\left(0, 1-\frac{t_i}{T_i}\right)
    \end{equation}
    where $t_i \in \mathbb{R}_{\geq0}$ is the elapsed time from system entry to hospital arrival and $T_i \in \mathbb{R}_{>0}$ is the estimated maximum survival duration without intervention.
    
    \item Resource matching penalty $PQ_i \in [0, 1]$:
    \begin{equation}
    PQ_i = \frac{q_i}{Q_i}
    \end{equation}
    where $Q_i \in \mathbb{N}$ represents required medical resource types and $q_i \in \{0, 1,...,Q_i\}$ represents available resource types at the assigned hospital.
\end{itemize}

The individual reward components $R_i$ are calculated based on patient severity status $s_i$ and hospital level $h \in \{1, 2, 3\}$: 

\begin{itemize}
    \item For newly deceased patients:
    \begin{equation}
    R_i = 
    \begin{cases}
    -600 & \text{if } s_i = 3 \text{ (critical)} \\
    -400 & \text{if } s_i = 2 \text{ (severe)}
    \end{cases}
    \end{equation}
    
    \item For critical patients ($s_i = 3$):
    \begin{equation}
    R_i = 300PQ_i + 
    \begin{cases}
    300PT_i & \text{if } h = 1 \\
    150PT_i & \text{if } h = 2 \\
    0 & \text{if } h = 3
    \end{cases}
    \end{equation}
    
    \item For severe patients ($s_i = 2$):
    \begin{equation}
    R_i = 200PQ_i + 
    \begin{cases}
    200PT_i & \text{if } h = 1 \\
    200PT_i & \text{if } h = 2 \\
    100PT_i & \text{if } h = 3
    \end{cases}
    \end{equation}
    
    \item For minor patients ($s_i = 1$):
    \begin{equation}
    R_i = 100PQ_i + 
    \begin{cases}
    0 & \text{if } h = 1 \\
    50PT_i & \text{if } h = 2 \\
    100PT_i & \text{if } h = 3
    \end{cases}
    \end{equation}
    
    \item For patients who expire after hospital assignment:
    \begin{equation}
    R_i = 
    \begin{cases}
    -300 & \text{if } h = 1 \\
    -200 & \text{if } h = 2 \\
    -100 & \text{if } h = 3
    \end{cases}
    \end{equation}
\end{itemize}

This reward structure is justified by several key principles in MCI management:

\begin{itemize}
    \item \textbf{Resource-Capability Optimization}: The model balances hospital levels with specific medical capabilities. While prioritizing higher-level hospitals for critical cases, it rewards best-effort allocations such as level 2 hospitals with appropriate medical capabilities, ensuring efficient utilization of the entire healthcare network's resources.
    
    \item \textbf{Survival Window Optimization}: The reward structure dynamically adjusts rewards based on the critical time-to-treatment window, emphasizing the importance of rapid intervention particularly for time-sensitive conditions.
    
    \item \textbf{Mortality Risk Minimization}: The significant negative rewards associated with patient mortality guides the model toward decisions that maximize survival probability, particularly in cases requiring time-critical interventions.
    
    \item \textbf{Adaptable Decision Making}: The reward structure accommodates real-world constraints by recognizing that optimal care might be achieved through various hospital-level combinations when appropriate medical capabilities are present, rather than strictly adhering to trauma level hierarchies.
\end{itemize}

This comprehensive reward design ensures the model learns to balance multiple competing objectives in MCI patient distribution: appropriate level of care, resource availability, survival probability, and system-wide efficiency.

\subsection{MasTER: The Simulation Platform}\label{MasTER}

MasTER (Mass-Casualty Trauma and Emergency Response) is an intelligent human-in-the-loop command dashboard accessible as a web application that simulates MCIs and provides end-users with a virtual environment to make transfer decisions. On the platform, users can assess the various trauma patients, injury mechanism, injury severity classification according to accepted standards for MCI, as well as the potential hospitals they can transfer to, including travel time and the various resources at each site (e.g. availability of intensive care unit beds, operating rooms, ventilators, blood products, etc.). The platform was designed to be capable of simulating MCIs in any geographic location, any combination of total number of trauma victims, injury mechanism/severity, available institutions and resources at each site. To optimize the fidelity of the simulation environment, the identification of trauma victims, and availability of transportation vehicles and hospital resources follows a sigmoid relationship as time progresses. Users can assign patients to be transferred to specific hospitals either through direct drag-and-drop function or by requesting AI-generated suggestions, which can be accepted or declined.

The user interface of MasTER has 6 major components, in the sequence of numbered components in Figure \ref{figure1-top}: 1) \textit{Jurisdiction Selector} enables hierarchical navigation of geographical responsibility areas with radius-based refinement; 2) \textit{Notifications} delivers color-coded, timestamped updates on system events for quick assessment; 3) \textit{Status Bar} provides real-time metrics including elapsed time, unassigned patients by severity, mortality count, and available ambulances; 4) \textit{Detail Panel} displays comprehensive information about selected patients (severity, injuries, resource needs) or hospitals (level, capabilities, distance); 5) \textit{Interactive Map} visualizes incident site and hospitals with multiple viewing options for improved spatial awareness.; 6) \textit{Draggable Action Panel} presents two synchronized lists: unassigned patients and hospitals in the selected jurisdiction by default. Patients are color-coded by severity (Critical in red, Severe in yellow, Minor in green, and Deceased in gray), while hospitals are differentiated by their trauma level (Level 1-3). Users can assign patients either through direct drag-and-drop interactions or by requesting AI-generated suggestions, which can be accepted or declined; an example of AI suggestion is given in Figure \ref{figure1-bottom}. We organized Table \ref{tab:component-challenges} showing the exact mapping relationships between each component and the challenges stated in Section \ref{Introduction}.

\section{Data Availability}

We generated simulated MCI environments specifically for DRL model training. No datasets were used nor produced, as the model training relied on agent interaction with procedurally generated environments rather than offline training.

\section{Code Availability}

The model was trained utilizing the open-source Gymnasium library \cite{gymnasium} from OpenAI and the Stable-Baselines3 framework \cite{stable-baselines3}. The web-accessible dashboard was developed with React framework \cite{react} for the front-end interface, Google Maps API \cite{googlemaps} for geospatial visualization, and integrated with UHN's secure database backend APIs hosted on Microsoft Azure infrastructure \cite{azure}. 

The complete source code repository of MasTER, including the training and testing of the DRL model, and the web-accessible dashboard, is available from the corresponding author upon reasonable request.

\newpage
\bibliography{main}

\begin{thebibliography}{10}
\urlstyle{rm}
\expandafter\ifx\csname url\endcsname\relax
  \def\url#1{\texttt{#1}}\fi
\expandafter\ifx\csname urlprefix\endcsname\relax\def\urlprefix{URL }\fi
\expandafter\ifx\csname doiprefix\endcsname\relax\def\doiprefix{DOI: }\fi
\providecommand{\bibinfo}[2]{#2}
\providecommand{\eprint}[2][]{\url{#2}}

\bibitem{bazyar2019triage}
\bibinfo{author}{Bazyar, J.}, \bibinfo{author}{Farrokhi, M.} \& \bibinfo{author}{Khankeh, H.}
\newblock \bibinfo{journal}{\bibinfo{title}{Triage systems in mass casualty incidents and disasters: a review study with a worldwide approach}}.
\newblock {\emph{\JournalTitle{Open access Macedonian journal of medical sciences}}} \textbf{\bibinfo{volume}{7}}, \bibinfo{pages}{482} (\bibinfo{year}{2019}).

\bibitem{bolduc2018comparison}
\bibinfo{author}{Bolduc, C.}, \bibinfo{author}{Maghraby, N.}, \bibinfo{author}{Fok, P.}, \bibinfo{author}{Luong, T.-L.~M.} \& \bibinfo{author}{Homier, V.}
\newblock \bibinfo{journal}{\bibinfo{title}{Comparison of electronic versus manual mass-casualty incident triage}}.
\newblock {\emph{\JournalTitle{Prehospital and Disaster Medicine}}} \textbf{\bibinfo{volume}{33}}, \bibinfo{pages}{273--278} (\bibinfo{year}{2018}).

\bibitem{tahernejad2024application}
\bibinfo{author}{Tahernejad, A.}, \bibinfo{author}{Sahebi, A.}, \bibinfo{author}{Abadi, A. S.~S.} \& \bibinfo{author}{Safari, M.}
\newblock \bibinfo{journal}{\bibinfo{title}{Application of artificial intelligence in triage in emergencies and disasters: a systematic review}}.
\newblock {\emph{\JournalTitle{BMC Public Health}}} \textbf{\bibinfo{volume}{24}}, \bibinfo{pages}{3203} (\bibinfo{year}{2024}).

\bibitem{albahri2024systematic}
\bibinfo{author}{Albahri, A.} \emph{et~al.}
\newblock \bibinfo{journal}{\bibinfo{title}{A systematic review of trustworthy artificial intelligence applications in natural disasters}}.
\newblock {\emph{\JournalTitle{Computers and Electrical Engineering}}} \textbf{\bibinfo{volume}{118}}, \bibinfo{pages}{109409} (\bibinfo{year}{2024}).

\bibitem{bazyar2022accuracy}
\bibinfo{author}{Bazyar, J.}, \bibinfo{author}{Farrokhi, M.}, \bibinfo{author}{Salari, A.}, \bibinfo{author}{Safarpour, H.} \& \bibinfo{author}{Khankeh, H.~R.}
\newblock \bibinfo{journal}{\bibinfo{title}{Accuracy of triage systems in disasters and mass casualty incidents; a systematic review}}.
\newblock {\emph{\JournalTitle{Archives of academic emergency medicine}}} \textbf{\bibinfo{volume}{10}} (\bibinfo{year}{2022}).

\bibitem{khorram2023implication}
\bibinfo{author}{Khorram-Manesh, A.} \emph{et~al.}
\newblock \bibinfo{journal}{\bibinfo{title}{The implication of a translational triage tool in mass casualty incidents: part three: a multinational study, using validated patient cards}}.
\newblock {\emph{\JournalTitle{Scandinavian Journal of Trauma, Resuscitation and Emergency Medicine}}} \textbf{\bibinfo{volume}{31}}, \bibinfo{pages}{88} (\bibinfo{year}{2023}).

\bibitem{reinforcement-survey}
\bibinfo{author}{Shakya, A.~K.}, \bibinfo{author}{Pillai, G.} \& \bibinfo{author}{Chakrabarty, S.}
\newblock \bibinfo{journal}{\bibinfo{title}{Reinforcement learning algorithms: A brief survey}}.
\newblock {\emph{\JournalTitle{Expert Systems with Applications}}} \textbf{\bibinfo{volume}{231}}, \bibinfo{pages}{120495}, \doiprefix\url{https://doi.org/10.1016/j.eswa.2023.120495} (\bibinfo{year}{2023}).

\bibitem{sutton-reinforcement}
\bibinfo{author}{Sutton, R.~S.} \& \bibinfo{author}{Barto, A.~G.}
\newblock \emph{\bibinfo{title}{Reinforcement Learning: An Introduction}} (\bibinfo{publisher}{A Bradford Book}, \bibinfo{address}{Cambridge, MA, USA}, \bibinfo{year}{2018}).

\bibitem{LeCun-deep-learning}
\bibinfo{author}{LeCun, Y.}, \bibinfo{author}{Bengio, Y.} \& \bibinfo{author}{Hinton, G.}
\newblock \bibinfo{journal}{\bibinfo{title}{Deep learning}}.
\newblock {\emph{\JournalTitle{Nature}}} \textbf{\bibinfo{volume}{521}}, \bibinfo{pages}{436--444}, \doiprefix\url{10.1038/nature14539} (\bibinfo{year}{2015}).

\bibitem{ian-deep}
\bibinfo{author}{Goodfellow, I.}, \bibinfo{author}{Bengio, Y.} \& \bibinfo{author}{Courville, A.}
\newblock \emph{\bibinfo{title}{Deep Learning}} (\bibinfo{publisher}{The MIT Press}, \bibinfo{year}{2016}).

\bibitem{drl-survey}
\bibinfo{author}{Arulkumaran, K.}, \bibinfo{author}{Deisenroth, M.~P.}, \bibinfo{author}{Brundage, M.} \& \bibinfo{author}{Bharath, A.~A.}
\newblock \bibinfo{journal}{\bibinfo{title}{Deep reinforcement learning: A brief survey}}.
\newblock {\emph{\JournalTitle{IEEE Signal Processing Magazine}}} \textbf{\bibinfo{volume}{34}}, \bibinfo{pages}{26--38}, \doiprefix\url{10.1109/MSP.2017.2743240} (\bibinfo{year}{2017}).

\bibitem{mnih2016asynchronous}
\bibinfo{author}{Mnih, V.}
\newblock \bibinfo{title}{Asynchronous methods for deep reinforcement learning} (\bibinfo{year}{2016}).

\bibitem{chen2022syntheticdata}
\bibinfo{author}{Chen, A.} \& \bibinfo{author}{Chen, D.}
\newblock \bibinfo{journal}{\bibinfo{title}{Simulation of a machine learning enabled learning health system for risk prediction using synthetic patient data}}.
\newblock {\emph{\JournalTitle{Scientific Reports}}} \textbf{\bibinfo{volume}{12}}, \bibinfo{pages}{17917}, \doiprefix\url{10.1038/s41598-022-23011-4} (\bibinfo{year}{2022}).

\bibitem{baucum2021improvingdrl}
\bibinfo{author}{Baucum, M.}, \bibinfo{author}{Khojandi, A.} \& \bibinfo{author}{Vasudevan, R.}
\newblock \bibinfo{journal}{\bibinfo{title}{Improving deep reinforcement learning with transitional variational autoencoders: A healthcare application}}.
\newblock {\emph{\JournalTitle{IEEE Journal of Biomedical and Health Informatics}}} \textbf{\bibinfo{volume}{25}}, \bibinfo{pages}{2273--2280}, \doiprefix\url{10.1109/JBHI.2020.3027443} (\bibinfo{year}{2021}).

\bibitem{schulman2017proximal}
\bibinfo{author}{Schulman, J.}, \bibinfo{author}{Wolski, F.}, \bibinfo{author}{Dhariwal, P.}, \bibinfo{author}{Radford, A.} \& \bibinfo{author}{Klimov, O.}
\newblock \bibinfo{journal}{\bibinfo{title}{Proximal policy optimization algorithms}}.
\newblock {\emph{\JournalTitle{arXiv preprint arXiv:1707.06347}}}  (\bibinfo{year}{2017}).

\bibitem{mousavi2018deep}
\bibinfo{author}{Mousavi, S.}, \bibinfo{author}{Schukat, M.} \& \bibinfo{author}{Howley, E.}
\newblock \bibinfo{title}{Deep reinforcement learning: An overview}.
\newblock \bibinfo{pages}{426--440}, \doiprefix\url{10.1007/978-3-319-56991-8_32} (\bibinfo{year}{2018}).

\bibitem{yu2019reinforcement}
\bibinfo{author}{Yu, C.}, \bibinfo{author}{Liu, J.} \& \bibinfo{author}{Nemati, S.}
\newblock \bibinfo{journal}{\bibinfo{title}{Reinforcement learning in healthcare: A survey}}.
\newblock {\emph{\JournalTitle{ACM Computing Surveys (CSUR)}}} \textbf{\bibinfo{volume}{52}}, \bibinfo{pages}{1--38} (\bibinfo{year}{2019}).

\bibitem{gandhi2020applications}
\bibinfo{author}{Gandhi, N.} \& \bibinfo{author}{Mishra, S.}
\newblock \bibinfo{journal}{\bibinfo{title}{Applications of reinforcement learning for medical decision making}}.
\newblock {\emph{\JournalTitle{CEUR Workshop Proceedings}}} \textbf{\bibinfo{volume}{2872}} (\bibinfo{year}{2020}).

\bibitem{greco2021artificial}
\bibinfo{author}{Greco, M.}, \bibinfo{author}{Caruso, P.~F.} \& \bibinfo{author}{Cecconi, M.}
\newblock \bibinfo{title}{Artificial intelligence in the intensive care unit}.
\newblock In \emph{\bibinfo{booktitle}{Seminars in respiratory and critical care medicine}}, vol.~\bibinfo{volume}{42}, \bibinfo{pages}{002--009} (\bibinfo{organization}{Thieme Medical Publishers, Inc.}, \bibinfo{year}{2021}).

\bibitem{ji2019deep}
\bibinfo{author}{Ji, S.}, \bibinfo{author}{Zheng, Y.}, \bibinfo{author}{Wang, Z.} \& \bibinfo{author}{Li, T.}
\newblock \bibinfo{journal}{\bibinfo{title}{A deep reinforcement learning-enabled dynamic redeployment system for mobile ambulances}}.
\newblock {\emph{\JournalTitle{Proceedings of the ACM on Interactive, Mobile, Wearable and Ubiquitous Technologies}}} \textbf{\bibinfo{volume}{3}}, \bibinfo{pages}{1--20} (\bibinfo{year}{2019}).

\bibitem{prasad2017reinforcement}
\bibinfo{author}{Prasad, N.}, \bibinfo{author}{Cheng, L.-F.}, \bibinfo{author}{Chivers, C.}, \bibinfo{author}{Draugelis, M.} \& \bibinfo{author}{Engelhardt, B.~E.}
\newblock \bibinfo{journal}{\bibinfo{title}{A reinforcement learning approach to weaning of mechanical ventilation in intensive care units}}.
\newblock {\emph{\JournalTitle{arXiv preprint arXiv:1704.06300}}}  (\bibinfo{year}{2017}).

\bibitem{lee2020improving}
\bibinfo{author}{Lee, S.} \& \bibinfo{author}{Lee, Y.~H.}
\newblock \bibinfo{title}{Improving emergency department efficiency by patient scheduling using deep reinforcement learning}.
\newblock In \emph{\bibinfo{booktitle}{Healthcare}}, vol.~\bibinfo{volume}{8}, \bibinfo{pages}{77} (\bibinfo{organization}{MDPI}, \bibinfo{year}{2020}).

\bibitem{liu2018deep}
\bibinfo{author}{Liu, S.} \& \bibinfo{author}{See, K.~C.}
\newblock \bibinfo{title}{Deep reinforcement learning for clinical decision support: A brief survey}.
\newblock In \emph{\bibinfo{booktitle}{AMIA Annual Symposium Proceedings}}, vol. \bibinfo{volume}{2018}, \bibinfo{pages}{1442} (\bibinfo{organization}{American Medical Informatics Association}, \bibinfo{year}{2018}).

\bibitem{mahmood2018benchmarking}
\bibinfo{author}{Mahmood, A.~R.}, \bibinfo{author}{Korenkevych, D.}, \bibinfo{author}{Vasan, G.}, \bibinfo{author}{Ma, W.} \& \bibinfo{author}{Bergstra, J.}
\newblock \bibinfo{journal}{\bibinfo{title}{Benchmarking reinforcement learning algorithms on real-world robots}}.
\newblock {\emph{\JournalTitle{arXiv preprint arXiv:1809.07731}}}  (\bibinfo{year}{2018}).

\bibitem{zhavoronkov2019deep}
\bibinfo{author}{Zhavoronkov, A.} \emph{et~al.}
\newblock \bibinfo{journal}{\bibinfo{title}{Deep learning enables rapid identification of potent ddr1 kinase inhibitors}}.
\newblock {\emph{\JournalTitle{Nature biotechnology}}} \textbf{\bibinfo{volume}{37}}, \bibinfo{pages}{1038--1040} (\bibinfo{year}{2019}).

\bibitem{yauney2018reinforcement}
\bibinfo{author}{Yauney, G.} \& \bibinfo{author}{Shah, P.}
\newblock \bibinfo{title}{Reinforcement learning with action-derived rewards for chemotherapy and clinical trial dosing regimen selection}.
\newblock In \emph{\bibinfo{booktitle}{Machine Learning for Healthcare Conference}}, \bibinfo{pages}{161--226} (\bibinfo{organization}{PMLR}, \bibinfo{year}{2018}).

\bibitem{hart1986nasa}
\bibinfo{author}{Hart, S.~G.} \& \bibinfo{author}{Staveland, L.~E.}
\newblock \bibinfo{title}{Development of nasa-tlx (task load index): Results of empirical and theoretical research}.
\newblock In \bibinfo{editor}{Hancock, P.~A.} \& \bibinfo{editor}{Meshkati, N.} (eds.) \emph{\bibinfo{booktitle}{Human Mental Workload}}, vol.~\bibinfo{volume}{52} of \emph{\bibinfo{series}{Advances in Psychology}}, \bibinfo{pages}{139--183}, \doiprefix\url{https://doi.org/10.1016/S0166-4115(08)62386-9} (\bibinfo{publisher}{North-Holland}, \bibinfo{year}{1988}).

\bibitem{sus}
\bibinfo{author}{Brooke, J.}
\newblock \bibinfo{title}{Sus-a quick and dirty usability scale}.
\newblock In \emph{\bibinfo{booktitle}{Usability evaluation in industry}}, \bibinfo{pages}{189--194} (\bibinfo{publisher}{Taylor \& Francis}, \bibinfo{year}{1996}).

\bibitem{redcap1}
\bibinfo{author}{Harris, P.~A.} \emph{et~al.}
\newblock \bibinfo{journal}{\bibinfo{title}{Research electronic data capture (redcap)—a metadata-driven methodology and workflow process for providing translational research informatics support}}.
\newblock {\emph{\JournalTitle{Journal of Biomedical Informatics}}} \textbf{\bibinfo{volume}{42}}, \bibinfo{pages}{377--381} (\bibinfo{year}{2009}).

\bibitem{redcap2}
\bibinfo{author}{Harris, P.~A.} \emph{et~al.}
\newblock \bibinfo{journal}{\bibinfo{title}{The redcap consortium: Building an international community of software platform partners}}.
\newblock {\emph{\JournalTitle{Journal of Biomedical Informatics}}} \textbf{\bibinfo{volume}{95}}, \bibinfo{pages}{103208} (\bibinfo{year}{2019}).

\bibitem{nasatlx}
\bibinfo{author}{Hart, S.~G.} \& \bibinfo{author}{Staveland, L.~E.}
\newblock \bibinfo{journal}{\bibinfo{title}{Development of nasa-tlx (task load index): Results of empirical and theoretical research}}.
\newblock {\emph{\JournalTitle{Advances in Psychology}}} \textbf{\bibinfo{volume}{52}}, \bibinfo{pages}{139--183} (\bibinfo{year}{1988}).

\bibitem{clarkson2024ems}
\bibinfo{author}{Clarkson, L.} \& \bibinfo{author}{Williams, M.}
\newblock \emph{\bibinfo{title}{EMS Mass Casualty Triage}} (\bibinfo{publisher}{StatPearls Publishing}, \bibinfo{address}{Treasure Island (FL)}, \bibinfo{year}{2024}).
\newblock \bibinfo{note}{Updated 2023 Aug 8}.

\bibitem{gymnasium}
\bibinfo{author}{Towers, M.} \emph{et~al.}
\newblock \bibinfo{title}{Gymnasium: A standard {API} for reinforcement learning} (\bibinfo{year}{2023}).

\bibitem{stable-baselines3}
\bibinfo{author}{Raffin, A.} \emph{et~al.}
\newblock \bibinfo{title}{Stable-baselines3: Reliable reinforcement learning implementations}.
\newblock \bibinfo{howpublished}{\url{https://github.com/DLR-RM/stable-baselines3}} (\bibinfo{year}{2021}).

\bibitem{react}
\bibinfo{author}{{Meta Platforms, Inc.}}
\newblock \bibinfo{title}{React: A javascript library for building user interfaces}.
\newblock \bibinfo{howpublished}{\url{https://react.dev}} (\bibinfo{year}{2023}).
\newblock \bibinfo{note}{Accessed: 2024-03-01}.

\bibitem{googlemaps}
\bibinfo{author}{{Google LLC}}.
\newblock \bibinfo{title}{Google maps platform: Maps javascript api}.
\newblock \bibinfo{howpublished}{\url{https://developers.google.com/maps/documentation/javascript}} (\bibinfo{year}{2024}).
\newblock \bibinfo{note}{Accessed: 2024-03-01}.

\bibitem{azure}
\bibinfo{author}{Corporation, M.}
\newblock \bibinfo{title}{Microsoft azure: Cloud computing services}.
\newblock \bibinfo{howpublished}{Cloud Computing Platform} (\bibinfo{year}{2024}).

\end{thebibliography}

\end{document}